\documentclass[10pt,twocolumn,letterpaper]{article}

\usepackage{iccv}
\usepackage{times}
\usepackage{epsfig}
\usepackage{graphicx}
\usepackage{amsmath}
\usepackage{amssymb}

\usepackage{caption}
\usepackage{subcaption}
\usepackage{multirow}
\usepackage{bigstrut}
\usepackage{hhline}
\usepackage{gensymb}
\usepackage{booktabs}
\usepackage{mdframed}
\usepackage{color}
\usepackage{substr}
\usepackage{breqn}
\usepackage{mathtools}
\usepackage{xfrac}
\usepackage{balance}
\usepackage[normalem]{ulem}
\usepackage{tabularx}

\usepackage[pagebackref=true,breaklinks=true,letterpaper=true,colorlinks,bookmarks=false]{hyperref}

\usepackage{cleveref}

\iccvfinalcopy

\DeclarePairedDelimiter\floor{\lfloor}{\rfloor}

\begin{document}
\newcommand\tab[1][2.5cm]{\hspace*{#1}}
\newcommand\tabs[1][1.7cm]{\hspace*{#1}}

\title{Hybrid Learning of Optical Flow and Next Frame Prediction to Boost Optical Flow in the Wild}

\author{Nima Sedaghat, Mohammadreza Zolfaghari, Thomas Brox\\
University of Freiburg\\
Germany\\
{\tt\small \{nima,zolfagha,brox\}@cs.uni-freiburg.de}
}

\maketitle

\begin{abstract}
CNN-based optical flow estimation has attracted attention recently, mainly due to its impressively high frame rates.
These networks perform well on synthetic datasets, but they are still far behind the classical methods in real-world videos. This is because there is no ground truth optical flow for training these networks on real data. In this paper, we boost CNN-based optical flow estimation in real scenes with the help of the freely available self-supervised task of next-frame prediction. To this end, we train the network in a hybrid way, providing it with a mixture of synthetic and real videos. With the help of a sample-variant multi-tasking architecture, the network is trained on different tasks depending on the availability of ground-truth. We also experiment with the prediction of ``next-flow'' instead of estimation of the current flow, which is intuitively closer to the task of next-frame prediction and yields favorable results.
We demonstrate the improvement in optical flow estimation on the real-world KITTI benchmark. Additionally, we test the optical flow indirectly in an action classification scenario. As a side product of this work, we report significant improvements over state-of-the-art in the task of next-frame prediction.
\end{abstract}

\begin{figure}[]
  \begin{center}
    \includegraphics[width=\linewidth]{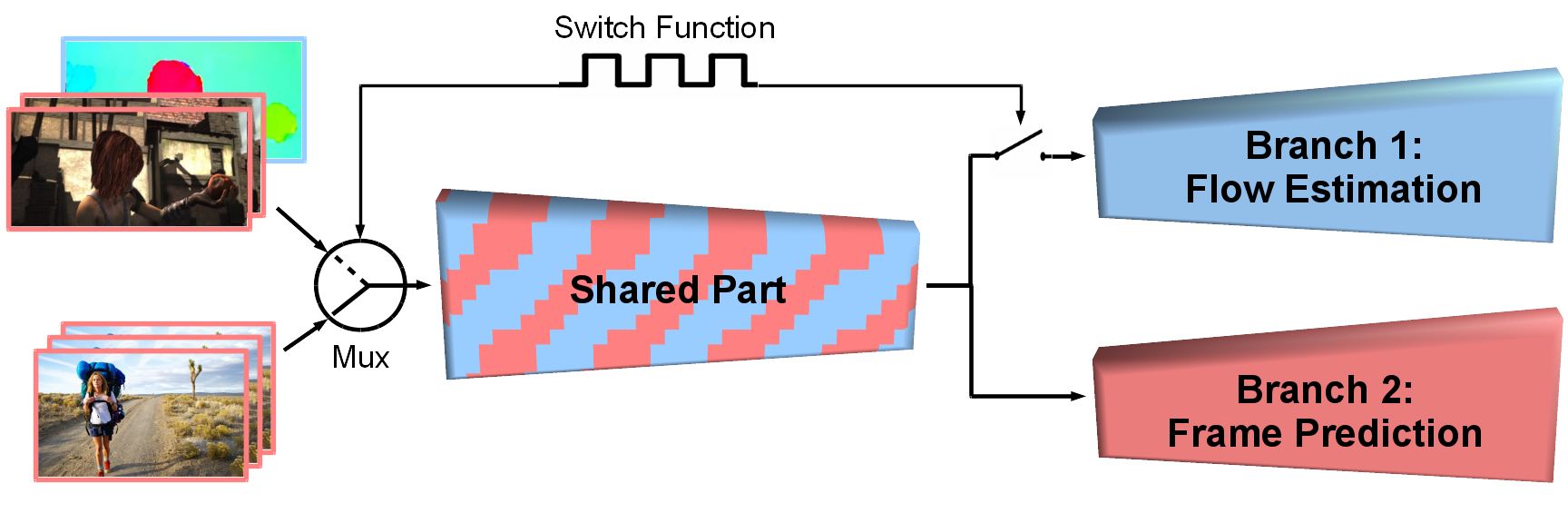}\vspace{-2mm}
  \end{center}
  \caption{We improve CNN-based optical flow estimation in real videos by adding the extra \emph{self-supervised} task of future frame prediction, and training the network with a mixture of synthetic and real-world videos. This combination is made possible by putting a ``multiplexer'' at the entry of the network which mixes data from the two sources on a timely basis. 
  }
  \label{fig:teaser}
\end{figure}

\begin{figure*}[]
  \begin{center}
    \includegraphics[width=1\linewidth]{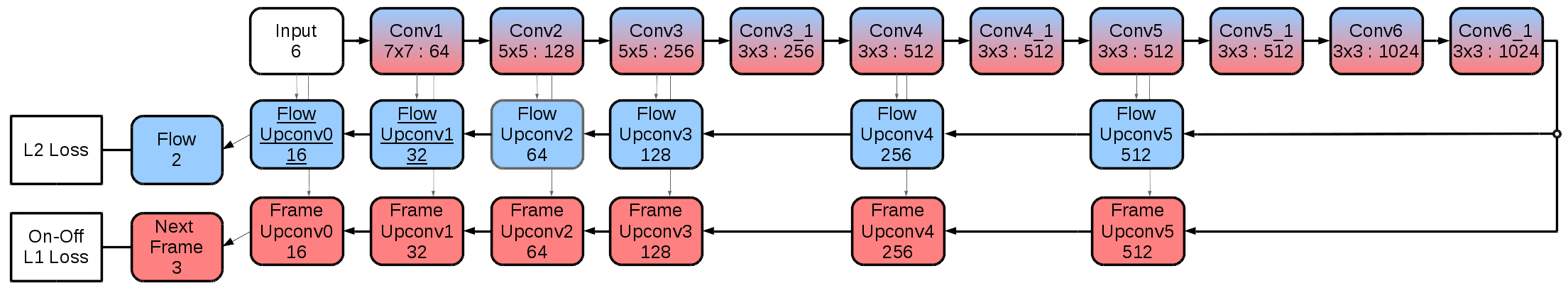}
  \end{center}
  \caption{Details of the multi-tasking architecture. For the sake of clarity, the lower-resolution outputs and their corresponding losses (as introduced in \cite{dosovitskiy_flownet:_2015}) are not displayed here. The flow estimation and frame prediction branches only differ in the number of channels of the output layer(s). Numbers in the boxes show kernel size and number of output channels for each layer. The upconvolutional layers are designed such that the output of each layer is of the same size (resolution) as its dual convolutional layer. As it is a fully convolutional architecture, the output resolution of each layer varies, depending on the resolution of the input frames.}
  \label{fig:architecture}
\end{figure*}

\section{Introduction}

Supervised learning of optical flow estimation with a deep network yields a good trade-off between run time and accuracy of the estimated optical flow \cite{dosovitskiy_flownet:_2015}. However, such supervised learning requires a large number of training pairs, which have been  provided via synthetic images. Such imagery lacks realism and diversity, and it keeps the network from using the full potential of the learning concept. Particularly on real-world data, FlowNet~\cite{dosovitskiy_flownet:_2015} does not yield the same accuracy as state-of-the-art conventional optical flow estimation techniques.

In this paper, we approach this problem by providing real-world data to the network during training. Since there is only a very limited amount of real-world image pairs with ground truth optical flow, we use a semi-supervised hybrid multi-tasking scheme that exploits real-world videos without ground truth and synthetic imagery with ground truth. For the network to learn useful concepts from the unlabeled data, we build on the self-supervised task of next-frame prediction as an auxiliary task. The general concept of this hybrid learning task is illustrated in \Cref{fig:teaser}. 

The hybrid multi-tasking combines the best of supervised learning on synthetic data and self-supervised learning on real data. On the KITTI optical flow benchmark, we obtained a clear improvement over the FlowNet, which was trained without the self-supervised next frame prediction task. The improvement over the baseline is even larger when testing on an application task for optical flow, such as action recognition.  

In addition to the hybrid multi-task learning of optical flow estimation and next frame prediction, we also propose multi-task learning on next frame prediction and next flow prediction. The latter two sub-tasks are more compatible and improve results when feeding the optical flow into an action recognition network. 

While we mainly focus on improving optical flow with the auxiliary task of next frame prediction, we also show benefits on next frame prediction.

\section{Related Work}
Since the work by Horn \& Schunk \cite{horn_determining_1981}, optical flow estimation has been dominated by variational methods \cite{brox_high_2004, perez2013tv, revaud_epicflow:_2015}.

The FlowNet by Dosovitskiy \etal~\cite{dosovitskiy_flownet:_2015} was the first deep network trained end-to-end on optical flow estimation. It was followed by Teney \etal~\cite{teney_learning_2016} and Tran \etal~\cite{tran_deep_2016}. These supervised learning methods require training data with optical flow annotations. In Dosovitskiy et al.~\cite{dosovitskiy_flownet:_2015} and Mayer et al.~\cite{mayer_large_2016} synthetic datasets were introduced to provide such data.
Tran \etal~\cite{tran_deep_2016} applied an existing variational method to create pseudo-ground truth data. 

Instead, Ahmadi and Patras~\cite{ahmadi_unsupervised_2016} and Yu \etal~\cite{yu_back_2016} formulated the task as an unsupervised learning problem. To this end, they used a cost function based on the classical color constancy assumption, as it is used in variational techniques.

Video prediction has been very popular recently \cite{mathieu_deep_2015,xue_visual_2016,finn_unsupervised_2016,lotter_deep_2016, jayaraman_look-ahead_2016,mahjourian_geometry-based_2016,saito_temporal_2016,patraucean_spatio-temporal_2015}. Although some of these works focus on prediction as the main objective \cite{mathieu_deep_2015,xue_visual_2016}, most of them use it as an auxiliary task. 
Finn \etal~\cite{finn_unsupervised_2016} proposed an action-conditioned video prediction model to facilitate unsupervised learning for physical interaction.
Patraucean \etal~\cite{patraucean_spatio-temporal_2015} learn optical flow by warping the current frame to the next one.
Lotter \etal~\cite{lotter_deep_2016} use prediction to learn representations for object recognition.

The works by Pintea \etal~\cite{pintea_deja_2014}, Walker \etal~\cite{walker_uncertain_2016,walker_dense_2015}, Jayaraman \etal~\cite{jayaraman_look-ahead_2016}, and Vondrick \etal~\cite{vondrick_anticipating_2016} focus on motion prediction. Their predicted motion is conditioned on a single input frame. In contrast,  we model future motion based on current motion and the scene content by making explicit use of two consecutive frames as input.

\section{Hybrid Architecture and Training Schedule}
\subsection{Optical Flow Estimation} \label{Flow Estimation}

The flow estimation network in the proposed hybrid architecture largely builds on the FlowNet architecture introduced in Dosovitskiy \etal~\cite{dosovitskiy_flownet:_2015}. 
As illustrated in \Cref{fig:architecture}, we add two more up-convolutional layers (Upconv1, Upconv0) to the decoder. This yields a flow field with the resolution of the input images. This is advantageous when combining the network with next-frame prediction. In contrast, the network in Dosovitskiy et al.~\cite{dosovitskiy_flownet:_2015} yields a lower resolution flow field, which is up-sampled with bilinear interpolation. 

In \Cref{fig:architecture}, the first and second row compose the encoder and decoder components of the flow estimator respectively. While our network follows the same multi-resolution scheme as in Dosovitskiy et al.~\cite{dosovitskiy_flownet:_2015}, in \Cref{fig:architecture} we omit the extra details regarding the so-called refinement steps (Figure 3 of \cite{dosovitskiy_flownet:_2015}) which represent lower-resolution outputs. We use the endpoint error loss (EPE) for training of this branch of the network. A more detailed illustration of the architecture is provided in the supplementary material.

\subsection{Next-Frame Prediction} \label{Next-Frame Prediction}

The network for the auxiliary task of next-frame prediction shares the encoder with the flow estimation network, and adds a second decoder stream with independent weights but using the same architecture. Rows 1 \& 3 in \Cref{fig:architecture} form the next-frame prediction component of the network. 
As suggested in previous work \cite{mathieu_deep_2015}, we use an L1 loss to avoid blur in the generated images. 

As reported in \cref{Experiments}, we experimented with different number of frames as input for next-frame prediction. However, in the multi-tasking scheme, we only use a 2-frame set-up to be compatible with the paired task of flow estimation. The two three-channel RGB images are provided as a stacked six-channel input to the overall network.

\subsection{Joint training} \label{multi-tasking}

For joint training of the hybrid network, there are two challenges. First, the data comes from two different sources, and there are multiple ways how to mix them during training. Secondly, unlike synthetic data, the real data does not come with optical flow ground truth, i.e., for real data as input, there is no loss for the flow related stream of the network.

\paragraph{Hybrid data}
We mix the data at the minibatch level: data in a single batch is taken completely either from the synthetic dataset or the real-world dataset. The minibatch $\mathcal{B}^{(i)}$ at iteration $i$ alternates between minibatches $\mathcal{B}_1^{(i)}$ \& $\mathcal{B}_2^{(i)}$ from the two data sources
\begin{equation} \label{eq:mixture}
  \mathcal{B}^{(i)} = (1-s^{(i)}) \mathcal{B}_1^{(i)} + s^{(i)} \mathcal{B}_2^{(i)}
\end{equation}
using the \emph{switch} function
\begin{equation} \label{eq:switch_function}
  s^{(i)} = \floor{ \frac{i\mod{(n_1+n_2)}}{n_1} },
\end{equation}
which always yields 0 or 1 and allows for different numbers of cycles $n_1$ and $n_2$ dedicated to each data source, respectively. 

\paragraph{Batch-variant loss}
The total loss at the $i^{th}$ iteration is computed according to:
\begin{equation} \label{eq:time_variant_loss}
  \mathcal{L}^{(i)} = w_1\mathcal{L}_1^{(i)}s^{(i)} + w_2\mathcal{L}_2^{(i)}
\end{equation}
in which $\mathcal{L}_1^{(i)}$ \&  $\mathcal{L}_2^{(i)}$ are the flow and frame estimation losses respectively, with their assigned weights, $w_1$ \& $w_2$. 

In case of real-data without ground truth optical flow, we deactivate the flow related loss $\mathcal{L}_1^{(i)}$ and, thus, the flow related decoder stream of the network. Both the loss and the loss gradient are set to zero. 
We keep the loss in sync with the switch function $s$ to ensure the desired functionality: the network learns on both tasks when synthetic data is provided, but skips updating the optical flow decoder when there is no ground truth. 

We set the loss weights such that $w_1/w_2$ is equal to $\sigma_2/\sigma_1$ where $\sigma$'s are the estimates of the variances of the input data (frame vs. flow) and are computed over a subset of 500 random samples from the training sets. In the following experiments, we report results with a fixed ratio of $1/5$ for $w_{flow}/w_{frame}$.

In our main experiments we fix the ratio of cycles dedicated to synthetic and real data sources. But we also provide an analysis on the effect of different cycle ratios on the quality of the output flow field.

\begin{figure}
  \begin{center}
    \begin{subfigure}[b]{\linewidth} \centering
      \includegraphics[width=.85\textwidth]{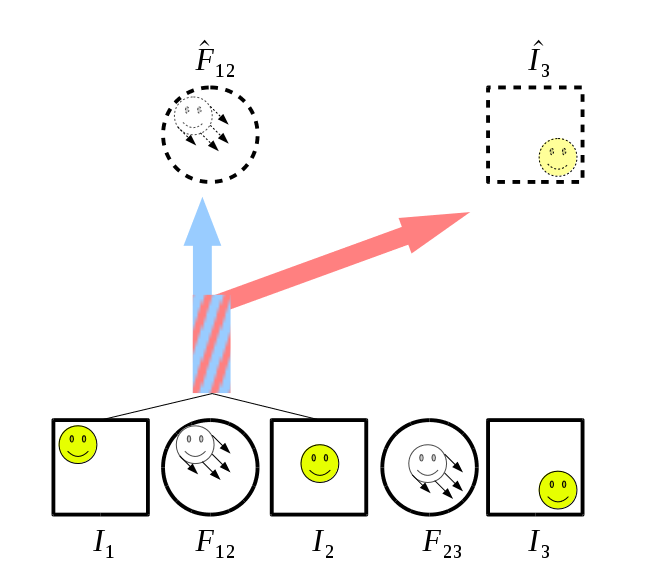}
      \caption{}\label{fig:next_flow_a}
    \end{subfigure}
    \begin{subfigure}[b]{\linewidth} \centering
      \includegraphics[width=.85\textwidth]{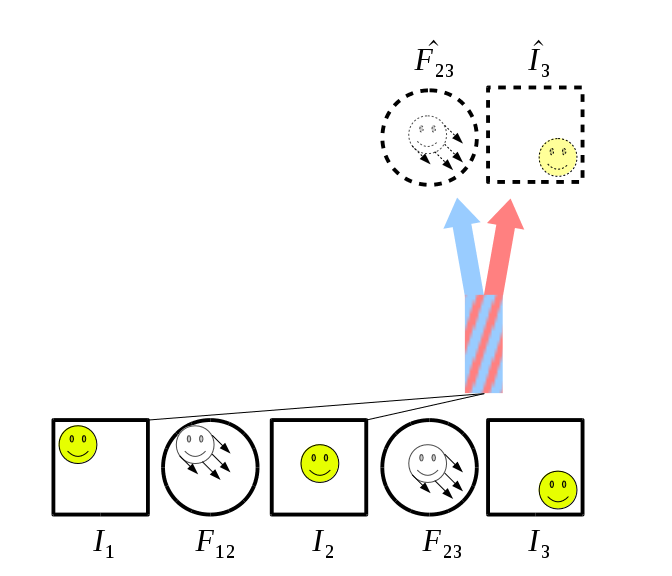}
      \caption{}\label{fig:next_flow_b}\vspace{-5mm}
    \end{subfigure}
  \end{center}
  \caption{Illustration of the two multi-tasking schemes introduced in this paper. (a) Combination of flow estimation with next-frame prediction; (b) the next-flow prediction replaces the task of current flow estimation. Each $I_k$ denotes a single frame in a video sequence of length 3, and $F$ denotes a flow field. In both scenarios only $I_1$ and $I_2$ are the inputs to the network. Therefore, the terms ``current flow'' and ``next-flow'' refer to $F_{12}$ and $F_{23}$, respectively.}
  \label{fig:next_flow_diagram}
\end{figure}

\begin{figure*}[!ht]
        \begin{center}
        \newcolumntype{Y}{>{\small\centering\arraybackslash}X}
        \begin{tabularx}{\textwidth}{YYYYY}
            {Overlayed} & {FlowNet} & {FlowNet}     & {NextFlow} & {EpicFlow}\\
            {Inputs}    & {}        & {+NextFrame}  & {+NextFrame} & {}\\
        \end{tabularx}
        \end{center}
  \begin{center}
    \includegraphics[width=\linewidth]{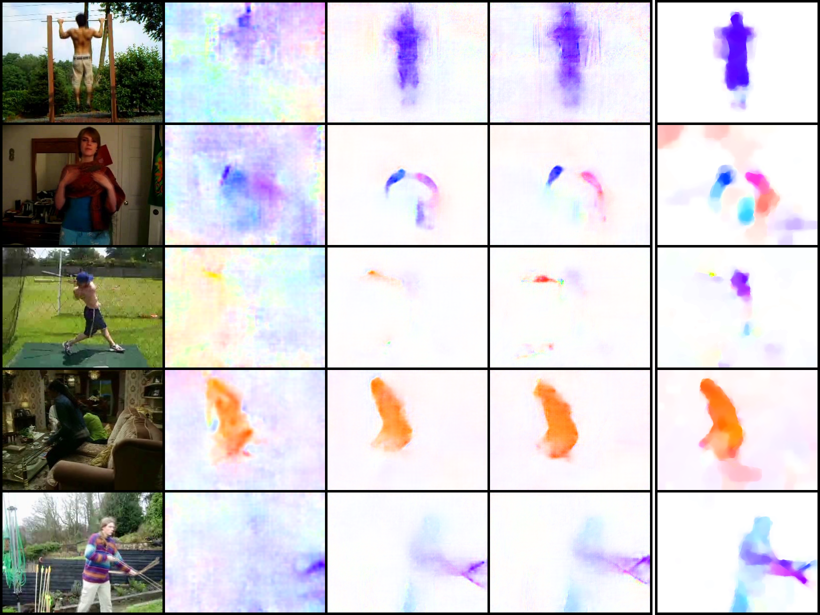}\vspace{-3mm}
  \end{center}
  \caption{Some samples from the estimated optical flow fields on real scenes from HMDB51~\cite{jhuang2011large}. Both of our suggested methods in the middle columns, show clear improvements in the flow fields and preserving the object shapes.}
  \label{fig:qualitative}
\end{figure*}

\begin{figure}
  \begin{center}
    \begin{subfigure}[b]{\linewidth} \centering
      \caption{First input frame}\label{fig:kitti_input}\vspace{-2mm}
      \includegraphics[width=1\textwidth]{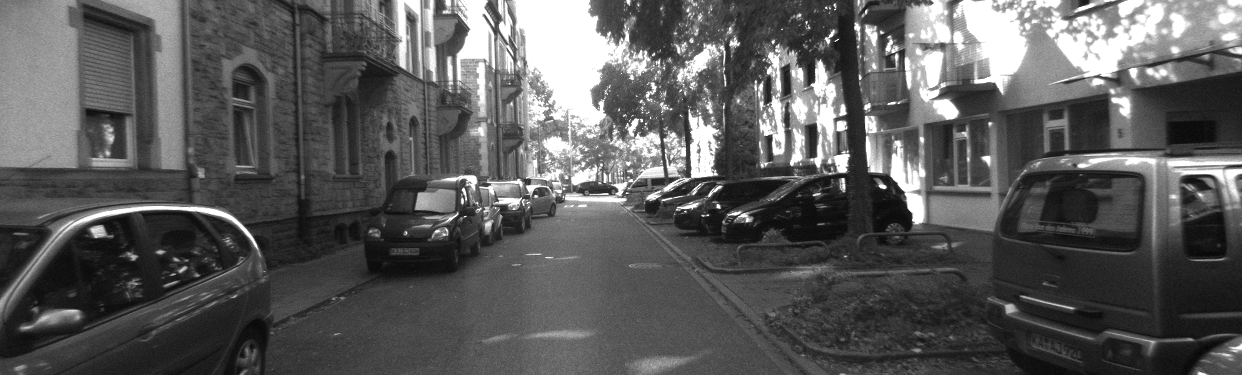}
    \end{subfigure}
    \begin{subfigure}[b]{\linewidth} \centering
      \caption{FlowNetS+ft~\cite{dosovitskiy_flownet:_2015}}\label{fig:kitti_a}\vspace{-2mm}
      \includegraphics[width=1\textwidth]{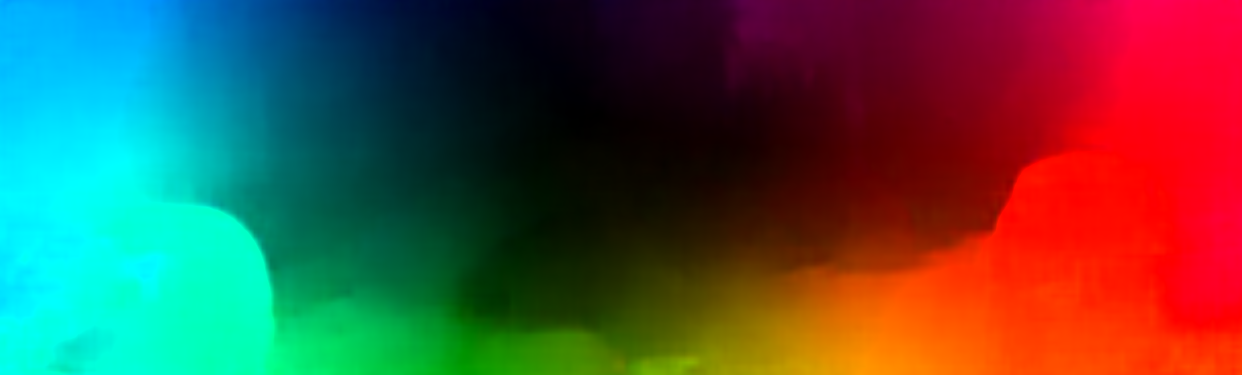}
    \end{subfigure}
    \begin{subfigure}[b]{\linewidth} \centering
      \caption{B2B Unsupervised FlowNet~\cite{yu_back_2016}}\label{fig:kitti_b}\vspace{-2mm}
      \includegraphics[width=1\textwidth]{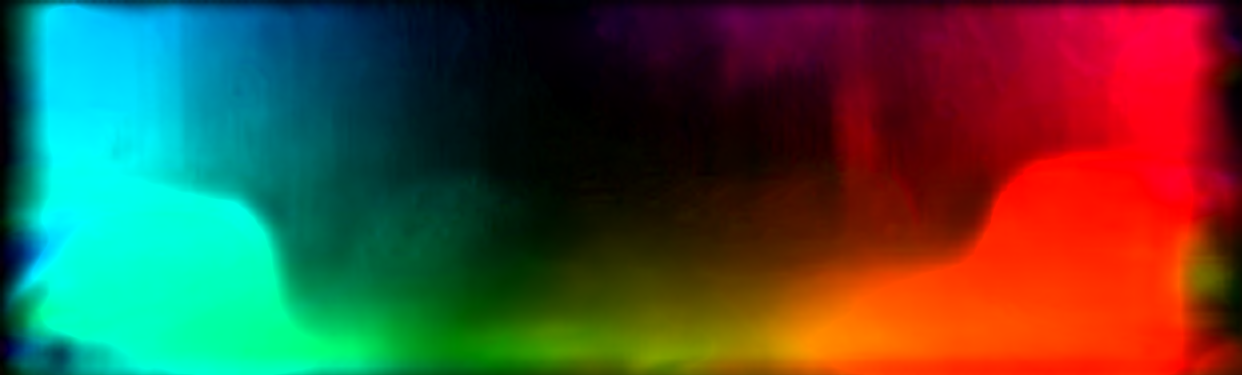}
    \end{subfigure}
    \begin{subfigure}[b]{\linewidth} \centering
      \caption{Ours}\label{fig:kitti_c}\vspace{-2mm}
      \includegraphics[width=1\textwidth]{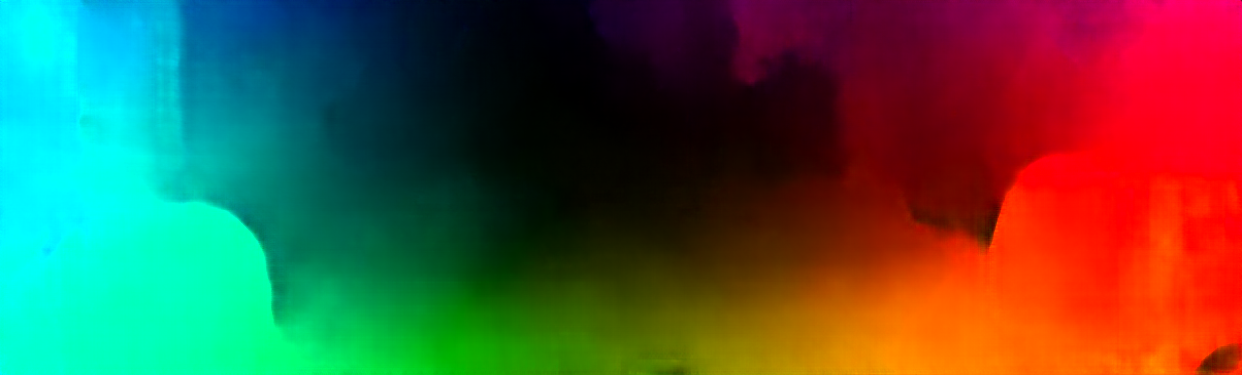}
    \end{subfigure}
  \end{center}
  \caption{Sample result on the KITTI benchmark. The unsupervised method of Yu \etal \cite{yu_back_2016} has problems near the image boundaries and reveals blurred motion boundaries.
  Our method shows a similar quality as FlowNetS+ft, although it has been only fine-tuned on unlabeled data.}
  \label{fig:kitti}
\end{figure}
\subsection{Next-Flow Prediction} \label{Next-Flow Prediction}

In a multi-tasking scheme, for the combination to yield significant improvements in the results, the two tasks need to be ``related'' \cite{caruana_multitask_1998}. 
In context of the current work, we hypothesize that prediction of the ``next-flow'' (i.e. the future flow to come), may have more in common with the task of next-frame prediction.
Figure \ref{fig:next_flow_diagram} compares the two combinations and gives an intuition on how the two ``prediction'' tasks match.

More formally: The two tasks share the encoder component of the network, that maps $(I_1,I_2) \mapsto z$, where $z$ is the internal representation to be learned by the network. This mapping is affected by both tasks during the backward pass. For multi-tasking to make sense, we expect that some features learned by the encoder are beneficial for both target tasks.
We hypothesize that the pair of $(z \mapsto I_3 , z \mapsto F_{23})$ have more to share, compared to $(z \mapsto I_3,z \mapsto F_{12})$ not only due to both being ``prediction'' tasks, but also because in obtaining the future flow, the network needs to learn, at least implicitly, about the future frame. This is not the case for the current flow (Figure \ref{fig:next_flow_diagram}). This hypothesis is supported by our experimental results.

\subsection{Training Details} \label{Training}
We train the network for 1 million iterations, with a batch
size of 8 for both of the data sources. The initial learning rate is 0.0001, and drops by a factor of 0.5 every 100K iterations starting from 300K. We use ADAM \cite{kingma_adam:_2014} for optimization with $\beta_1 = 0.9$, $\beta_2 = 0.999$. On an NVIDIA Titan~X, training takes roughly 10 days.

\renewcommand{\multirowsetup}{\centering} 
\setlength{\tabcolsep}{4pt}
\begin{table*}[t]
\small
  \begin{center}
    \begin{tabular}{ll|ccccccccc}
      \toprule

      {} & \multicolumn{1}{c}{Dataset for} & \multicolumn{2}{c}{KITTI'12} & {} & \multicolumn{1}{c}{KITTI'15} &
            & \multicolumn{1}{c}{Sintel} & {} & {FlyingThings3D}\\
      \cmidrule{3-4} \cmidrule{6-6}  \cmidrule{8-8}  \cmidrule{10-10}
      {} & \multicolumn{1}{c}{Frame Prediction\textbf{}} & {train} & {test} & {} & {train} & {} & {train} & {} & {test} \\
      \midrule

      B2B Unsupervised FlowNet \cite{yu_back_2016}&-                 &-          & 11.3   && -              && -   && -\\
      FlowNet \cite{dosovitskiy_flownet:_2015}& -                        &8.26          & -     && -              && 4.50  && -\\
      \midrule							    	                                         
      FlowNet Baseline    & -                                        &8.79          & -     && 15.59          && 4.33 && 1.84\\
      FlowNet Baseline+NextFrame   & Sports                                   &8.55          & -     && 15.16          && 4.38  && 1.86\\
      FlowNet Baseline+NextFrame   & Cityscapes                               &8.49          & -     && 14.68          && 4.24  && 1.80\\
      FlowNet Baseline+NextFrame   & KITTI:frames                             &8.37          & -     && 14.15          && 4.30   && 1.84   \\
      FlowNet Baseline+NextFrame   & Sports + KITTI:frames                    &8.39          & -     && 15.08              && 4.29   && 1.86   \\
      FlowNet Baseline+NextFrame   & Sports$\rightarrow$KITTI:frames          &\textbf{7.78} & \textbf{9.2}&& \textbf{13.95} && 4.36  && 1.85\\
      \midrule							    	                                         
      FlowNet \cite{dosovitskiy_flownet:_2015} $\rightarrow$KITTI:flow  & -	                     & 7.52         & 9.1   && -              && -       && -   \\
      FlowNet+NextFrame $\rightarrow$KITTI:flow   & KITTI:frames     & \textbf{5.31}& -     && \textbf{10.19}&& 5.35      && 2.82   \\

      \bottomrule
    \end{tabular}\vspace{-5mm}
  \end{center}
  \caption{Quantitative evaluation of optical flow estimation performance based on End Point Error (EPE). ``KITTI:frames'' indicates video frames (without flow annotations) from the KITTI dataset. Moreover, wherever the evaluation is performed on a KITTI (2012/2015) training subset, the data used for the training of the network is taken from its counterpart (2015/2012). The $\rightarrow$ sign indicates a pre-training/fine-tuning process. }
  \label{table:epe}
\end{table*}

\section{Experiments} \label{Experiments}

\subsection{Datasets}
For hybrid training of the network we need 2 datasets per experiment. We used the so-called ``FlyingThings3D'' dataset of Mayer \etal.~\cite{mayer_large_2016} as the data source with ground truth optical flow. It consists of more than 20000 training images and allows training a network from scratch. Moreover, it provides an independent test set that we used for testing. 
The much smaller Sintel dataset \cite{Butler:ECCV:2012} has 1064 samples and was used only for testing. 

The only available real-world dataset with ground truth optical flow is the KITTI dataset. There are two independent datasets, KITTI 2012~\cite{Geiger2012CVPR} and KITTI 2015~\cite{Menze2015CVPR}. We used both datasets for the quantitative evaluation of the optical flow. Since both datasets are independent, we always used one for training and the other for testing. Except for one experiment, we did not use the optical flow ground truth for training but only the images.  We took the frames from the ``multi-view'' extension of the datasets, consisting of 4074 and 4200 images in the 2012 and 2015 versions, respectively.

There are many large real-world datasets without optical flow ground truth. We used mainly a subset of the Sports1M dataset \cite{karpathy_large-scale_2014} for the self-supervised training task. The subset includes all videos with a file size up to 5 MBytes, amounting to more than 220K videos and 220M frames. We will make the selection list available online. Also in another experiment, we simply used 50000 frames from videos of the Cityscapes dataset \cite{cordts_cityscapes_2016}.

Moreover, we used the UCF101 \cite{soomro_ucf101:_2012} and HMDB51 \cite{jhuang2011large} datasets for testing the optical flow indirectly in an action recognition scenario. The datasets contain more than 2M \& 600K frames, respectively. 

To compare the performance of our next-frame predictor to published work, we used the same subset of UCF101 \cite{soomro_ucf101:_2012} as Mathieu et al.~\cite{mathieu_deep_2015}. It consists of 387 videos.

\subsection{Direct Evaluation of the Optical Flow}
In Figure \ref{fig:qualitative} we visualize some of the flow fields estimated with our method, FlowNet \cite{dosovitskiy_flownet:_2015}, and an accurate but slow variational method (EpicFlow \cite{revaud_epicflow:_2015}). On real-world scenes, our flow fields capture the shape of moving objects much better than the baseline FlowNet. We believe this sharpness is a result of asking for pixel-level accurate results in the auxiliary task of frame prediction, which regulates the blurring that the flow branch tends to exert.

We quantitatively evaluated the method on KITTI 2012 \& 2015. \Cref{table:epe} shows these results along with two synthetic datasets. All the experiments used the same synthetic source of data and they differ only in the source of real data.
`FlowNet Baseline' is our full-resolution extension of the architecture of \cite{dosovitskiy_flownet:_2015} trained on FlyingThings3D. 
`FlowNet+NextFrame` indicates our hybrid multi-tasking scheme. 

Results from various configurations are displayed in \Cref{table:epe}. Although the Sports dataset has little similarity with the scenes in KITTI, using this data for the auxiliary task yields significant improvements on KITTI. Using frames from the Cityscapes dataset, improves the results even more, as the videos are recorded in a similar context to that of KITTI's.
There is no significant change on the synthetic datasets. This does not come as a surprise, since the FlyingThings3D dataset can cover other synthetic datasets like Sintel well. There is no significant domain shift from the training set to the test set in this case.

\renewcommand{\multirowsetup}{\centering} 
\setlength{\tabcolsep}{4pt}
\begin{table*}[t]
\small
  \begin{center}
    \begin{tabular}{clcc}
      \toprule

      {} & {} & \multicolumn{2}{c}{Action Accuracy (\%)} \\
        \cmidrule{3-4}

      {} & {} & {UCF101 \cite{soomro_ucf101:_2012}} & {HMDB51 \cite{jhuang2011large}}\\

      \midrule

      \multirow{2}{*}{Classical}
      &EpicFlow \cite{revaud_epicflow:_2015}                            & 82.8 & 56.1\\
      &TV-L1 \cite{perez2013tv} (as reported in \cite{wang2016temporal})& 87.2 & -\\
      \cmidrule{1-4}
      \multirow{2}{*}{CNN based}
      &FlowNet                                                          & 62.0 		& 38.6\\
      &FlowNet pre-trained with NextFrame 		                        & 63.4		& 38.4\\
      &FlowNet+NextFrame Multi-tasking (1:5)	                        & 74.1 		& 48.4\\
      &NextFlow+NextFrame Multi-tasking	(1:5)	                        & \textbf{75.5}	& \textbf{48.9}\\
      \bottomrule

    \end{tabular}\vspace{-5mm}
  \end{center}
  \caption{Action classification accuracy. Each row contains results of training and testing the action classifier on optical flow generated by a specific method. 1:5 indicates the real to synthetic iterations ratio.}
  \label{table:action}
\end{table*}

\renewcommand{\multirowsetup}{\centering} 
\setlength{\tabcolsep}{4pt}
\begin{table}[]
\small
  \begin{center}
    \begin{tabular}{ccccccc}
      \toprule

      {} & \multicolumn{1}{c}{EPE} & {} & \multicolumn{2}{c}{Action Accuracy (\%)} \\

      \cmidrule{2-2} \cmidrule{4-5}
      {${n_{real}}:{n_{synth}}$} & {KITTI'12} & {} & {HMDB51} & {UCF101}\\
      \midrule

      FlowNet   & 8.88          && 38.6           & 62.0\\
      1:9		& 8.76          && 48.0	 	      & 75.3\\
      1:5		& \textbf{8.55} && \textbf{48.4}  & 74.1\\
      1:3		& 8.78          && 47.3           & 74.7\\
      1:1		& 8.94          && 48.3 		  & \textbf{76.6}\\
      4:1		& 10.35         && 48.0 		  & -\\

      \bottomrule
    \end{tabular}\vspace{-5mm}
  \end{center}
  \caption{Analysis of the effect of different cycles ratios on optical flow quality. For EPE, lower values are better. For action class accuracy, higher numbers are better.}
  \label{table:cycles_epe}
\end{table}
Using video frames from the KITTI dataset (labeled as `KITTI:frames') rather than the Sports or Cityscapes datasets for the auxiliary task, improves results on KITTI as expected.
We also experimented with combining the two real datasets, both in a parallel fashion and in a pre-training/fine-tuning scheme ('Sports$\rightarrow$KITTI:frames'). The latter led to another large improvement. We submitted this version to the official KITTI evaluation site to obtain results on the KITTI test set.  The result is essentially as good as the FlowNet fine-tuned on KITTI. Figure~\ref{fig:kitti} depicts a qualitative comparison on this benchmark.

We also report results on the fine-tuned FlowNet combined with the hybrid learning on the auxiliary task at the very bottom of \Cref{table:epe}. This experiment shows that even when fine-tuning the FlowNet baseline on KITTI, hybrid training still yields significant improvements. 

\newcommand\tabb[1][1.9cm]{\hspace*{#1}}
\newcommand\tabbs[1][.7cm]{\hspace*{#1}}
\begin{figure*}[t]
        \begin{center}
        \newcolumntype{Y}{>{\small\centering\arraybackslash}X}
        \begin{tabularx}{\textwidth}{YYYYYYYY}
            {} & {FlowNet} & {1:9} & {1:5} & {1:3} & {1:1} & {4:1} & {EpicFlow}\\
        \end{tabularx}
        \end{center}
  \begin{center}
    \includegraphics[width=\linewidth]{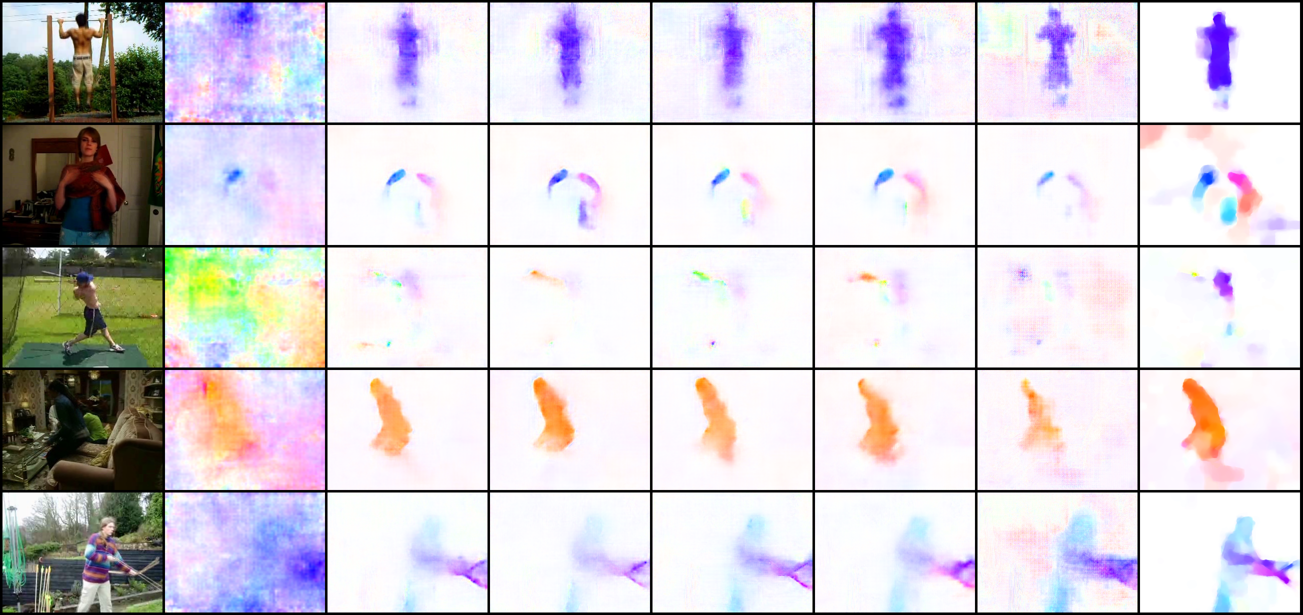}\vspace{-3mm}
  \end{center}
  \caption{Qualitative comparison of different data source combination cycles. On top of each column the $n_{real}:n_{synth}$ ratio is displayed. When there is no real data involved (FlowNet), the network fails to estimate an acceptable flow field in real scenes. On the other hand, if training spends too many cycles only on frame prediction, as in the 4:1 column, the network no longer focuses enough on the optical flow task. The best results are obtained with a ratio of 1:5 or 1:3.}
\label{fig:cycles}
\end{figure*}

\renewcommand{\multirowsetup}{\centering} 
\setlength{\tabcolsep}{4pt}
\begin{table*}[]
\small
  \begin{center}
    \begin{tabular}{cccccc}
      \toprule

      {} & {} & \multicolumn{2}{c}{Whole Image} & \multicolumn{2}{c}{Moving Regions} \\
      \cmidrule{3-4}
      \cmidrule{5-6}
      {} & {} & {Similarity}& {Sharpness}  & {Similarity}& {Sharpness} \\
      {} & {Method} & {PSNR(dB)}& {(dB)}  & {PSNR(dB)}& {(dB)} \\

      \midrule

      \multirow{3}{*}{Mathieu \etal \cite{mathieu_deep_2015}}
         & L1		& 22.3		& 18.5 		& 28.7 & 24.8\\
         & GDL + L1	& 23.9		& 18.7 		& 29.9 & 25\\
         & Adv + GDL + L1& 29.6		& 20.3 		& 32   & 25.4\\
      \cmidrule{1-6}
      Ours (2-frame)
         & L1		& 29.9		& 20.6 		& 31.9 & 25.4\\
      Ours (4-frame)
         & L1		& \textbf{30.8}	& \textbf{20.8} & 31.9 & 25.4\\

      \bottomrule
    \end{tabular}\vspace{-5mm}
  \end{center}
  \caption{Next frame prediction on UCF101 \cite{soomro_ucf101:_2012}. With just a simple L1 loss we already obtain clear improvements over the state-of-the-art.}
  \label{table:pred_results}
\end{table*}

\subsection{Indirect Evaluation: Action Classification}
As real-world videos rarely come with optical flow ground-truth (KITTI being an exception), possibilities for a direct evaluation of the optical flow is limited. Thus, we use the evaluation on flow-based action classification as an indirect quantitative measure on two larger real-world datasets. We use the action classifier network of Wang \etal~\cite{wang2016temporal} and train/test it with optical flow from different optical flow methods as input. 

Table \ref{table:action} shows the results of this evaluation. 
We used the Sports dataset to provide unsupervised data. The hybrid learning was done with a ratio of 1:5 for real to synthetic cycles.
The optical flow with our hybrid learning scheme improved results on action recognition by a large margin (12.1\% on UCF and 9.8\% on HMDB) when compared to the baseline FlowNet. We achieved even larger improvements by replacing current flow with `NextFlow'.

We also tried a pre-training/fine-tuning scenario in which the network is initially trained for the frame prediction task (on real data), and then fine-tuned with the main task (``FlowNet pre-trained with NextFrame''). Results confirm that this sequential learning is not sufficient. The multi-tasking scheme is necessary to make good use of the auxiliary task on the real data.

We report also number of two variational methods, TV-L1~\cite{zach_duality_2007} and EpicFlow~\cite{revaud_epicflow:_2015}. They provide a higher accuracy, but are also much slower than the network based approaches. 

\subsection{Impact of Task Combination Cycles}
We evaluated on which ratio of training cycles on synthetic and real data one obtains the best performance and on how robust the method is to deviations from the optimal ratio. 
We used the Sports dataset as data source in this experiment.
Figure \ref{fig:cycles} shows the results for various ${n_{real}}:{n_{synth}}$ cycle ratios.
Results are robust for a large range of ratios. Lower ratios approach the results of FlowNet, as the effect of the auxiliary task starts to vanish. Putting too much emphasis on the auxiliary task introduces artifacts in the optical flow field, since the network starts to care mostly about next frame prediction. In general, the ratio should be biased towards the supervised optical flow task. A ratio of 1:5 seems a good choice in general. 

\begin{figure*}[!ht]
        \begin{center}
        \newcolumntype{Y}{>{\small\centering\arraybackslash}X}
        \begin{tabularx}{\textwidth}{YYYYYYY}
            {$I_1$} & {$I_2$} & {$I_3$} & {$I_4$}  & {Ours} & {Mathieu \etal~\cite{mathieu_deep_2015}} & {Ground Truth}\\
        \end{tabularx}
        \end{center}
  \begin{center}
    \includegraphics[width=\linewidth]{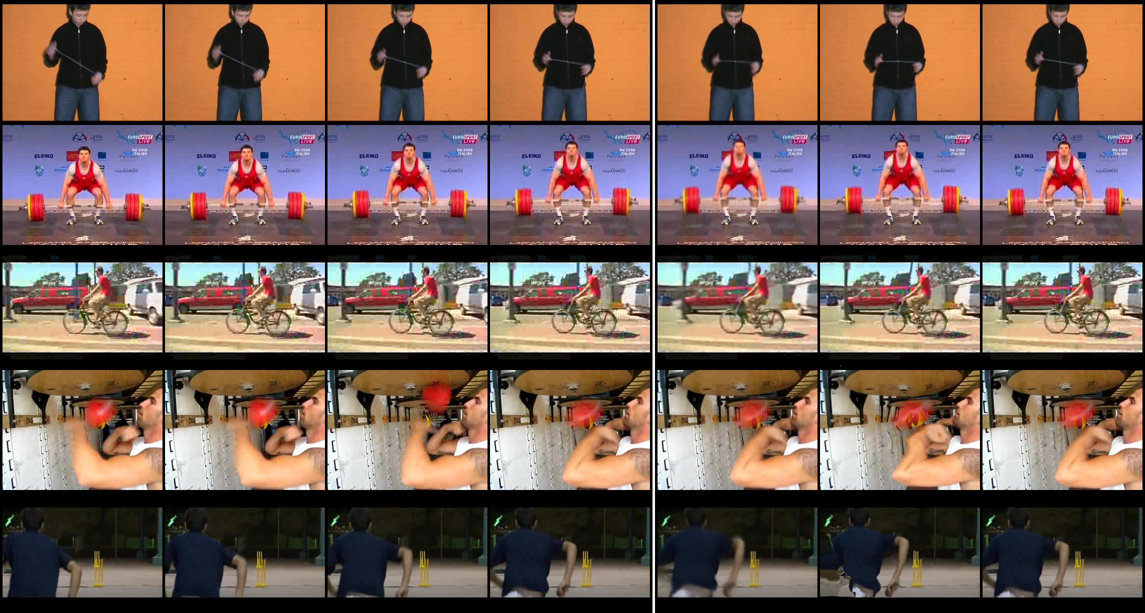}\vspace{-2mm}
  \end{center}
  \caption{Next frame prediction samples. Results of Mathieu \etal~\cite{mathieu_deep_2015} are often a bit sharper due to the adversarial loss, yet the method also introduces distortions and artifacts; see the last two samples. Our next frame predictions are blurrier due to relying only on the L1 loss, but yield robust predictions without distortion. This explains the on-par quantitative results in Table~\ref{table:pred_results}.}
  \label{fig:pred_res}
\end{figure*}

\subsection{Next-Frame Prediction as a Single Task}
We also evaluated the output of our next-frame prediction network and tested it on UCF. 
To this end, we trained it as an independent single-task network ($n_{synthetic}=0$). Table \ref{table:pred_results} shows a comparison with Mathieu \etal~\cite{mathieu_deep_2015} -- which to the best of our knowledge is the current state-of-the-art in next-frame prediction on UCF. Without the use of any auxiliary cost functions, as introduced in Mathieu \etal~\cite{mathieu_deep_2015} for the sake of sharp results, and just with a single L1 loss, we obtain results on par with Mathieu \etal on the moving regions of the image, and significantly better results on the whole image. This means that the network is more successful on applying the motion only to the dynamic areas and keeping the static areas intact.
We show qualitative examples of the predicted frames in Figure \ref{fig:pred_res} and in a video in the supplementary material.

Since frame prediction has only been an auxiliary task in the network, the input settings (particularly the cycles ratio) have been set to focus on improvement of the optical flow output. Therefore, by increasing the number of flow cycles, the next-frame prediction accuracy is degraded. 

\section{Conclusions}

We have presented a way to improve a deep network for optical flow estimation on real data by training it with an additional self-supervised auxiliary task. Our experiments showed a consistent improvement of the optical flow quality on real-world data. Thus, we believe that this approach largely improves the transfer of deep networks trained on synthetic dataset to domains in the real world. While we focused here on optical flow, the concept may transfer also to similar problems, such as disparity estimation, and alternative self-supervised auxiliary tasks. 

\section*{Acknowledgments}
We acknowledge funding by the ERC Starting Grant VideoLearn.

{\small
\bibliographystyle{ieee}
\bibliography{ref}

\begin{thebibliography}{10}\itemsep=-1pt

\bibitem{ahmadi_unsupervised_2016}
A.~Ahmadi and I.~Patras.
\newblock Unsupervised convolutional neural networks for motion estimation.
\newblock {\em arXiv:1601.06087 [cs]}, Jan. 2016.

\bibitem{brox_high_2004}
T.~Brox, A.~Bruhn, N.~Papenberg, and J.~Weickert.
\newblock High {{Accuracy Optical Flow Estimation Based}} on a {{Theory}} for
  {{Warping}}.
\newblock In {\em Computer {{Vision}} - {{ECCV}} 2004}, pages 25--36.
  {Springer, Berlin, Heidelberg}.

\bibitem{Butler:ECCV:2012}
D.~J. Butler, J.~Wulff, G.~B. Stanley, and M.~J. Black.
\newblock A naturalistic open source movie for optical flow evaluation.
\newblock In {A. Fitzgibbon et al. (Eds.)}, editor, {\em European {{Conf}}. on
  {{Computer Vision}} ({{ECCV}})}, Part IV, LNCS 7577, pages 611--625.
  {Springer-Verlag}, Oct. 2012.

\bibitem{caruana_multitask_1998}
R.~Caruana.
\newblock Multitask {{Learning}}.
\newblock In S.~Thrun and L.~Pratt, editors, {\em Learning to {{Learn}}}, pages
  95--133. {Springer US}, 1998.

\bibitem{cordts_cityscapes_2016}
M.~Cordts, M.~Omran, S.~Ramos, T.~Rehfeld, M.~Enzweiler, R.~Benenson,
  U.~Franke, S.~Roth, and B.~Schiele.
\newblock The {{Cityscapes Dataset}} for {{Semantic Urban Scene
  Understanding}}.
\newblock pages 3213--3223.

\bibitem{dosovitskiy_flownet:_2015}
A.~Dosovitskiy, P.~Fischer, E.~Ilg, P.~Hausser, C.~Hazirbas, V.~Golkov, P.~{van
  der Smagt}, D.~Cremers, and T.~Brox.
\newblock {{FlowNet}}: {{Learning Optical Flow With Convolutional Networks}}.
\newblock In {\em Proceedings of the {{IEEE International Conference}} on
  {{Computer Vision}}}, pages 2758--2766, 2015.

\bibitem{finn_unsupervised_2016}
C.~Finn, I.~Goodfellow, and S.~Levine.
\newblock Unsupervised learning for physical interaction through video
  prediction.
\newblock In {\em Advances {{In Neural Information Processing Systems}}}, pages
  64--72, 2016.

\bibitem{Geiger2012CVPR}
A.~Geiger, P.~Lenz, and R.~Urtasun.
\newblock Are we ready for {{Autonomous Driving}}? {{The KITTI Vision Benchmark
  Suite}}.
\newblock In {\em Conference on {{Computer Vision}} and {{Pattern Recognition}}
  ({{CVPR}})}.

\bibitem{horn_determining_1981}
B.~K.~P. Horn and B.~G. Schunck.
\newblock Determining optical flow.
\newblock {\em Artificial Intelligence}, 17(1):185--203, Aug. 1981.

\bibitem{jayaraman_look-ahead_2016}
D.~Jayaraman and K.~Grauman.
\newblock Look-{{Ahead Before You Leap}}: {{End}}-to-{{End Active Recognition}}
  by {{Forecasting}} the {{Effect}} of {{Motion}}.
\newblock In {\em Computer {{Vision}} \textendash{} {{ECCV}} 2016}, pages
  489--505. {Springer, Cham}, Oct. 2016.

\bibitem{jhuang2011large}
H.~Jhuang, H.~Garrote, E.~Poggio, T.~Serre, and T.~Hmdb.
\newblock A large video database for human motion recognition.
\newblock In {\em Proc. of {{IEEE International Conference}} on {{Computer
  Vision}}}, 2011.

\bibitem{karpathy_large-scale_2014}
A.~Karpathy, G.~Toderici, S.~Shetty, T.~Leung, R.~Sukthankar, and L.~Fei-Fei.
\newblock Large-scale {{Video Classification}} with {{Convolutional Neural
  Networks}}.
\newblock In {\em {{CVPR}}}, 2014.

\bibitem{kingma_adam:_2014}
D.~P. Kingma and J.~Ba.
\newblock Adam: {{A Method}} for {{Stochastic Optimization}}.

\bibitem{lotter_deep_2016}
W.~Lotter, G.~Kreiman, and D.~Cox.
\newblock Deep {{Predictive Coding Networks}} for {{Video Prediction}} and
  {{Unsupervised Learning}}.
\newblock {\em arXiv:1605.08104 [cs, q-bio]}, May 2016.

\bibitem{mahjourian_geometry-based_2016}
R.~Mahjourian, M.~Wicke, and A.~Angelova.
\newblock Geometry-{{Based Next Frame Prediction}} from {{Monocular Video}}.
\newblock {\em arXiv:1609.06377 [cs]}, Sept. 2016.

\bibitem{mathieu_deep_2015}
M.~Mathieu, C.~Couprie, and Y.~LeCun.
\newblock Deep multi-scale video prediction beyond mean square error.
\newblock {\em arXiv:1511.05440 [cs, stat]}, Nov. 2015.

\bibitem{mayer_large_2016}
N.~Mayer, E.~Ilg, P.~Hausser, P.~Fischer, D.~Cremers, A.~Dosovitskiy, and
  T.~Brox.
\newblock A large dataset to train convolutional networks for disparity,
  optical flow, and scene flow estimation.
\newblock In {\em Proceedings of the {{IEEE Conference}} on {{Computer Vision}}
  and {{Pattern Recognition}}}, pages 4040--4048, 2016.

\bibitem{Menze2015CVPR}
M.~Menze and A.~Geiger.
\newblock Object {{Scene Flow}} for {{Autonomous Vehicles}}.
\newblock In {\em Conference on {{Computer Vision}} and {{Pattern Recognition}}
  ({{CVPR}})}.

\bibitem{patraucean_spatio-temporal_2015}
V.~Patraucean, A.~Handa, and R.~Cipolla.
\newblock Spatio-temporal video autoencoder with differentiable memory.
\newblock {\em arXiv:1511.06309 [cs]}, Nov. 2015.

\bibitem{perez2013tv}
J.~S. P{\'e}rez, E.~Meinhardt-Llopis, and G.~Facciolo.
\newblock {{TV}}-{{L1}} optical flow estimation.
\newblock {\em Image Processing On Line}, 2013:137--150, 2013.

\bibitem{pintea_deja_2014}
S.~L. Pintea, J.~C. van Gemert, and A.~W.~M. Smeulders.
\newblock D{\'e}j{\`a} {{Vu}}:.
\newblock In D.~Fleet, T.~Pajdla, B.~Schiele, and T.~Tuytelaars, editors, {\em
  Computer {{Vision}} \textendash{} {{ECCV}} 2014}, number 8691 in Lecture
  Notes in Computer Science, pages 172--187. {Springer International
  Publishing}, Sept. 2014.

\bibitem{revaud_epicflow:_2015}
J.~Revaud, P.~Weinzaepfel, Z.~Harchaoui, and C.~Schmid.
\newblock {{EpicFlow}}: {{Edge}}-{{Preserving Interpolation}} of
  {{Correspondences}} for {{Optical Flow}}.
\newblock In {\em Proceedings of the {{IEEE}} Conference on Computer Vision and
  Pattern Recognition}, pages 1164--1172, 2015.

\bibitem{saito_temporal_2016}
M.~Saito and E.~Matsumoto.
\newblock Temporal {{Generative Adversarial Nets}}.
\newblock {\em arXiv:1611.06624 [cs]}, Nov. 2016.

\bibitem{soomro_ucf101:_2012}
K.~Soomro, A.~R. Zamir, and M.~Shah.
\newblock {{UCF101}}: {{A Dataset}} of 101 {{Human Actions Classes From
  Videos}} in {{The Wild}}.
\newblock {\em arXiv:1212.0402 [cs]}, Dec. 2012.

\bibitem{teney_learning_2016}
D.~Teney and M.~Hebert.
\newblock Learning to {{Extract Motion}} from {{Videos}} in {{Convolutional
  Neural Networks}}.
\newblock {\em arXiv:1601.07532 [cs]}, Jan. 2016.

\bibitem{tran_deep_2016}
D.~Tran, L.~Bourdev, R.~Fergus, L.~Torresani, and M.~Paluri.
\newblock Deep {{End2End Voxel2Voxel Prediction}}.
\newblock pages 17--24, 2016.

\bibitem{vondrick_anticipating_2016}
C.~Vondrick, H.~Pirsiavash, and A.~Torralba.
\newblock Anticipating {{Visual Representations From Unlabeled Video}}.
\newblock In {\em Proceedings of the {{IEEE Conference}} on {{Computer Vision}}
  and {{Pattern Recognition}}}, pages 98--106, 2016.

\bibitem{walker_uncertain_2016}
J.~Walker, C.~Doersch, A.~Gupta, and M.~Hebert.
\newblock An {{Uncertain Future}}: {{Forecasting}} from {{Static Images Using
  Variational Autoencoders}}.
\newblock In {\em Computer {{Vision}} \textendash{} {{ECCV}} 2016}, pages
  835--851. {Springer, Cham}, Oct. 2016.

\bibitem{walker_dense_2015}
J.~Walker, A.~Gupta, and M.~Hebert.
\newblock Dense {{Optical Flow Prediction}} from a {{Static Image}}.
\newblock {\em arXiv:1505.00295 [cs]}, May 2015.

\bibitem{wang2016temporal}
L.~Wang, Y.~Xiong, Z.~Wang, Y.~Qiao, D.~Lin, X.~Tang, and L.~Van~Gool.
\newblock Temporal segment networks: {{Towards}} good practices for deep action
  recognition.
\newblock In {\em European {{Conference}} on {{Computer Vision}}}, pages
  20--36. {Springer}, 2016.

\bibitem{xue_visual_2016}
T.~Xue, J.~Wu, K.~Bouman, and B.~Freeman.
\newblock Visual {{Dynamics}}: {{Probabilistic Future Frame Synthesis}} via
  {{Cross Convolutional Networks}}.
\newblock In D.~D. Lee, M.~Sugiyama, U.~V. Luxburg, I.~Guyon, and R.~Garnett,
  editors, {\em Advances in {{Neural Information Processing Systems}} 29},
  pages 91--99. {Curran Associates, Inc.}, 2016.

\bibitem{yu_back_2016}
J.~J. Yu, A.~W. Harley, and K.~G. Derpanis.
\newblock Back to {{Basics}}: {{Unsupervised Learning}} of {{Optical Flow}} via
  {{Brightness Constancy}} and {{Motion Smoothness}}.
\newblock {\em arXiv:1608.05842 [cs]}, Aug. 2016.

\bibitem{zach_duality_2007}
C.~Zach, T.~Pock, and H.~Bischof.
\newblock A {{Duality Based Approach}} for {{Realtime TV}}-{{L1 Optical Flow}}.
\newblock In {\em Pattern {{Recognition}}}, pages 214--223. {Springer, Berlin,
  Heidelberg}.

\end{thebibliography}
}

\newpage

\newlength\myheight
\newlength\mydepth
\settototalheight\myheight{Xygp}
\settodepth\mydepth{Xygp}
\setlength\fboxsep{0pt}
\newcommand*\inlinegraphics[1]{%
  \raisebox{-2pt}{\protect\includegraphics[height=10pt]{#1}}%
}

\begin{figure*}[t]
  \begin{center}
    \newcolumntype{Y}{>{\small\centering\arraybackslash}X}
    \centering{\Large{\textbf{Supplementary Material}}}
    \vspace*{2cm}
  \end{center}
  \begin{center}
    \includegraphics[width=1\linewidth]{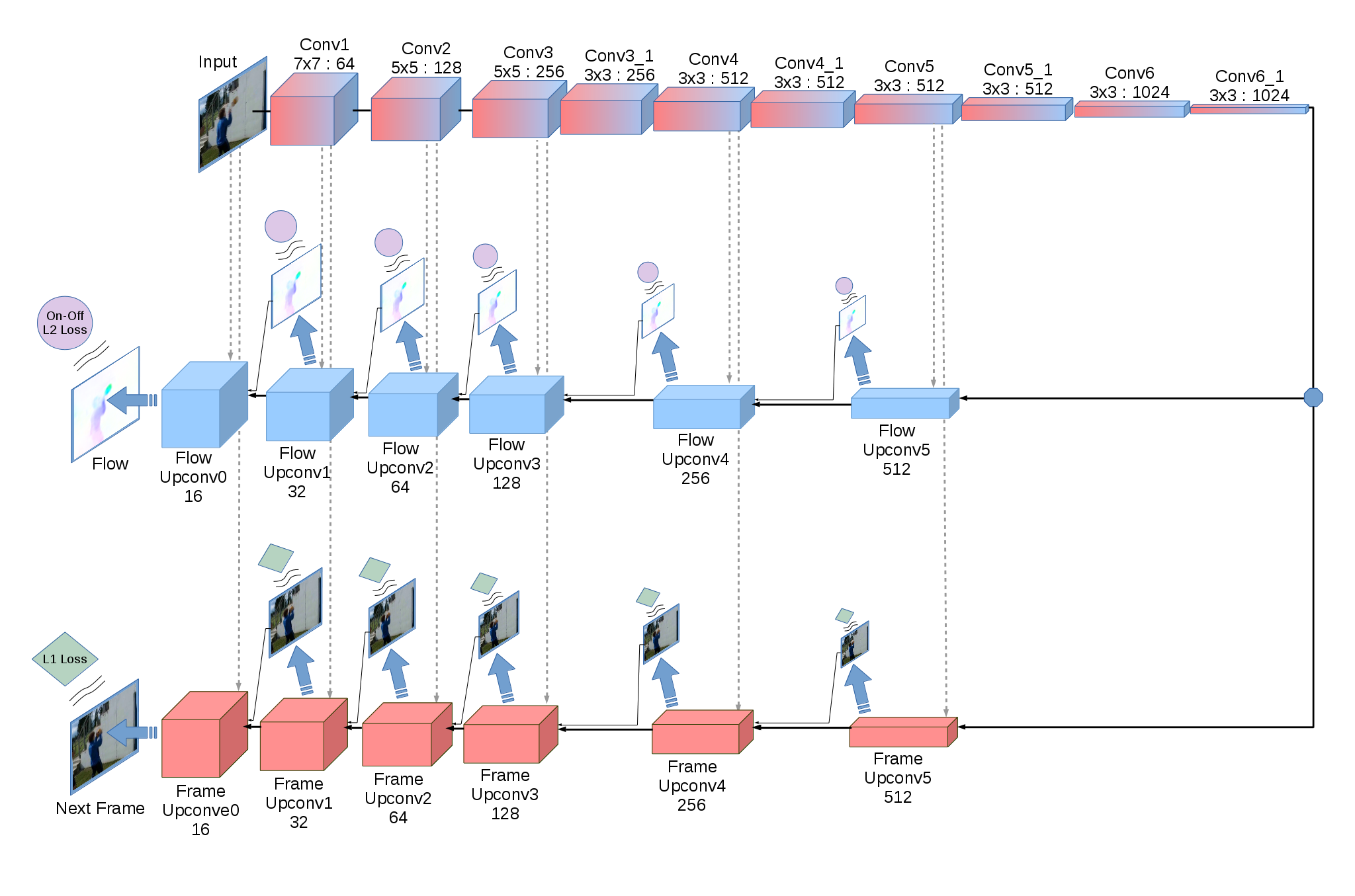}
  \end{center}
  \label{fig:architecture}
  \caption{
    In this detailed illustration of the architecture, we show the contraction throughout the encoder in the first row, followed by the two decoder branches, showing the expansion in the network. 
    Captions above/below the boxes show the layer names, as well as the number of outputs/feature maps.
    In the first row we also show the kernel sizes for each layer, while we do not display the fixed kernel size of 4x4 for the Upconv layers of rows 2 and 3 (see table~\ref{table:arch_details}). The fully-convolutional architecture can be used with different input sizes, and thus no resolution is displayed in this figure.
    Each \inlinegraphics{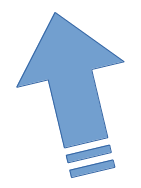} represents a convolutional layer with a kernel size of 3x3, and stride and padding values of 1, which preserves the spatial dimensions, and maps its higher dimensional input blob to a flow or frame prediction. These low resolution predictions are then up-sampled and concatenated to the input of the next layer, along with the corresponding features from the encoder. The ``On-Off'' losses of the flow prediction branch are all synchronized to (de)activate the branch when necessary.
    }
\end{figure*}

\renewcommand{\multirowsetup}{\centering} 
\setlength{\tabcolsep}{4pt}
\begin{table*}[t!]
\small
  \begin{center}
    \begin{tabular}{l|ccccccccccccc}
      \toprule
      {} & {Conv1} & {Conv2} & {Conv3} & {Conv3\_1} & {Conv4} & {Conv4\_1} & {Conv5} & {Conv5\_1} & {Conv6} & {Conv6\_1}\\
      \midrule
      kernel size & 7x7 & 5x5 & 5x5 & 3x3 & 3x3 & 3x3 & 3x3 & 3x3 & 3x3 & 3x3 \\
      stride & 2 & 2 & 2 & 1 & 2 & 1 & 2 & 1 & 2 & 1 \\
      padding & 3 & 2 & 2 & 1 & 1 & 1 & 1 & 1 & 1 & 1 \\
      \toprule
      {} & {Upconv5} & {Upconv4} & {Upconv3} & {Upconv2} & {Upconv1} & {Upconv0}\\
      \midrule
      kernel size & 4x4 & 4x4 & 4x4 & 4x4 & 4x4 & 4x4\\
      stride & 2 & 2 & 2 & 2 & 2 & 2 \\
      padding & 1 & 1 & 1 & 1 & 1 & 1 \\

      \bottomrule
    \end{tabular}\vspace{-5mm}
  \end{center}
  \caption{Kernel size, stride and padding settings for different layers of the network. Upconv layers in the flow and frame branches share the same settings.}
  \label{table:arch_details}
\end{table*}

\renewcommand{\multirowsetup}{\centering} 
\setlength{\tabcolsep}{4pt}
\begin{table*}[]
\small
  \begin{center}
    \begin{tabular}{c|cccccccc}
      \toprule
      {${n_{real}}:{n_{synth}}$} & {$\infty$} & {8:1} & {4:1} & {1:1} & {1:3} & {1:5} & {1:9}\\
      \midrule
      {Similarity - PSNR (dB) - Whole Image} & {29.9} & {29.20} & {29.14} & {28.8} & {28.57} & {28.34} & {28.24}\\

      \bottomrule
    \end{tabular}\vspace{-5mm}
  \end{center}
  \caption{Analysis of the effect of different cycles ratios on frame prediction. The more flow prediction cycles we include in the arrangement, the lower the prediction quality goes.}
  \label{table:pred_cycles}
  \vspace{12cm}
\end{table*}

\par\vspace{\fill}\pagebreak[0]\vspace{-\fill}

\end{document}